\documentclass[letterpaper,journal]{IEEEtran}
\usepackage{amsmath,amsfonts}
\usepackage{algorithmic}
\usepackage{array}
\usepackage[caption=false,font=normalsize,labelfont=sf,textfont=sf]{subfig}
\usepackage{textcomp}
\usepackage{stfloats}
\usepackage{url}
\usepackage{verbatim}
\usepackage{graphicx}
\usepackage{multirow}
\usepackage{booktabs} 
\usepackage{hyperref}
\def\BibTeX{{\rm B\kern-.05em{\sc i\kern-.025em b}\kern-.08em
    T\kern-.1667em\lower.7ex\hbox{E}\kern-.125emX}}
\usepackage{balance}
\begin{document}
\title{StereoDETR: Stereo-based Transformer for 3D Object Detection}
\author{Shiyi Mu, Zichong Gu, Zhiqi Ai, Anqi Liu, Yilin Gao and Shugong Xu*,~\IEEEmembership{Fellow,~IEEE}
\thanks{* Corresponding author

This work was supported in part by the 6G Science and Technology Innovation and Future Industry Cultivation Special Project of Shanghai Municipal Science and Technology Commission under Grant 24DP1501001, in part by  the National High Quality Program under Grant TC220H07D.

Shiyi Mu, Zichong Gu, Zhiqi Ai, Anqi Liu, and Yilin Gao are with Shanghai University, Shanghai 200444, China (e-mail: shiyimu@shu.edu.cn).

Shugong Xu is with Xi’an Jiaotong-Liverpool University, Suzhou 215123, China (e-mail: shugong.xu@xjtlu.edu.cn).

}}

\markboth{Journal of \LaTeX\ Class Files,~Vol.~18, No.~9, May~2025}%
{How to Use the IEEEtran \LaTeX \ Templates}

\IEEEpubid{\begin{minipage}{\textwidth}\ \\[12pt] \centering
		Copyright \copyright 20xx IEEE. Personal use of this material is permitted. Permission to use this material for any other purposes must \\ be obtained from the IEEE by sending an email to pubs-permissions@ieee.org.
\end{minipage}}

\maketitle

\begin{abstract}
 Compared to monocular 3D object detection, stereo-based 3D methods offer significantly higher accuracy but still suffer from high computational overhead and latency. The state-of-the-art stereo 3D detection method achieves twice the accuracy of monocular approaches, yet its inference speed is only half as fast. In this paper, we propose StereoDETR, an efficient stereo 3D object detection framework based on DETR. StereoDETR consists of two branches: a monocular DETR branch and a stereo branch. The DETR branch is built upon 2D DETR with additional channels for predicting object scale, orientation, and sampling points. The stereo branch leverages low-cost multi-scale disparity features to predict object-level depth maps. These two branches are coupled solely through a differentiable depth sampling strategy. To handle occlusion, we introduce a constrained supervision strategy for sampling points without requiring extra annotations. StereoDETR achieves real-time inference and is the first stereo-based method to surpass monocular approaches in speed. It also achieves competitive accuracy on the public KITTI benchmark, setting new state-of-the-art results on pedestrian and cyclist subsets. The code is available at https://github.com/shiyi-mu/StereoDETR-OPEN.
\end{abstract}


\section{Introduction}
\label{sec:intro}

With the widespread adoption of autonomous driving technologies, obstacle detection in open environments is facing increasingly stringent safety requirements. The task of 3D obstacle detection presents a fundamental trade-off between cost and accuracy. Existing solutions can be broadly categorized into LiDAR-based and vision-based approaches. LiDAR-based methods typically offer high accuracy but suffer from high sensor costs and substantial computational demands. In contrast, vision-based methods are more cost-effective and mainly include monocular and stereo approaches. Monocular vision methods\cite{monodetr_iccv23,Monodgp_cvpr25} are fast but often lack accuracy due to limited depth cues. Stereo vision methods\cite{YOLOStereo3D_2021,DSC3D_TCSVT25,TS3D_TITS24} improves depth estimation by leveraging disparity between dual-camera inputs, resulting in better accuracy. However, the additional computational burden often prevents stereo-based methods from achieving real-time performance, making them generally slower than monocular methods. 

\begin{figure}
    \centering
    \includegraphics[width=1\linewidth]{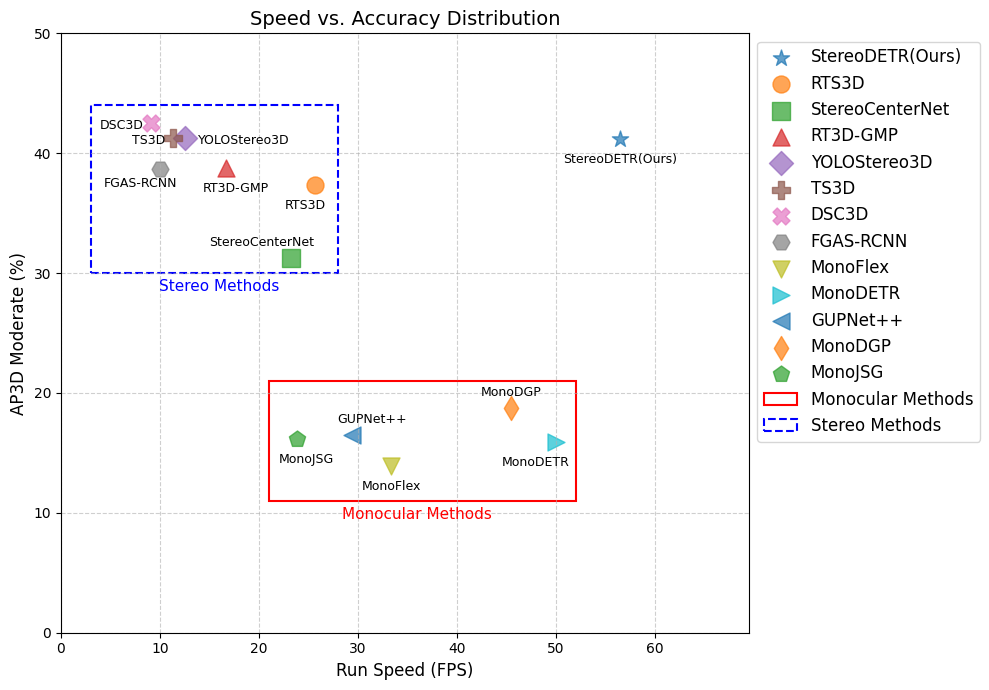}
    \caption{Comparison of accuracy and speed with existing camera-based methods on the KITTI test set (Car category, moderate difficulty).}

    \label{fig:ap_time}
\end{figure}

Monocular methods do not rely on point clouds or disparity estimation, making their architectures simple and efficient, with the fastest model\cite{monodetr_iccv23} running in as little as 20\,ms per frame. 
\IEEEpubidadjcol
The main challenge for monocular 3D detection lies in accurate depth estimation, often resulting in localization precision that is approximately half that of stereo-based methods\cite{DSC3D_TCSVT25}. Stereo-based approaches estimate depth through the disparity between stereo image pairs, offering higher 3D localization accuracy. However, disparity computation introduces significant latency. To reduce inference time, prior works have proposed region-of-interest (ROI)-based local matching\cite{Stereo-R-CNN-2019,TLNet_CVPR19,SIDE_WACV22} and low-resolution feature-level disparity estimation\cite{YOLOStereo3D_2021,DSC3D_TCSVT25,TS3D_TITS24}. Despite these efforts, the fastest stereo detector\cite{Rts3d_aaai21} still require around 39\,ms per frame. Fig~\ref{fig:ap_time} shows a comparison between more monocular and stereo 3D detectors in terms of speed and accuracy. The accuracy is reported as the AP$_{3D}$ for the \textit{Car} category on the KITTI benchmark under the moderate difficulty setting with an IoU threshold of 0.7. The horizontal axis denotes the runtime speed (FPS). Clearly, there exists a trade-off between accuracy and speed in existing monocular and stereo methods. Our proposed StereoDETR achieves both the accuracy of stereo-based models and a speed that surpasses existing monocular approaches.

\begin{figure*}[ht] % 双栏图环境
    \centering
    \includegraphics[width=0.95\linewidth]{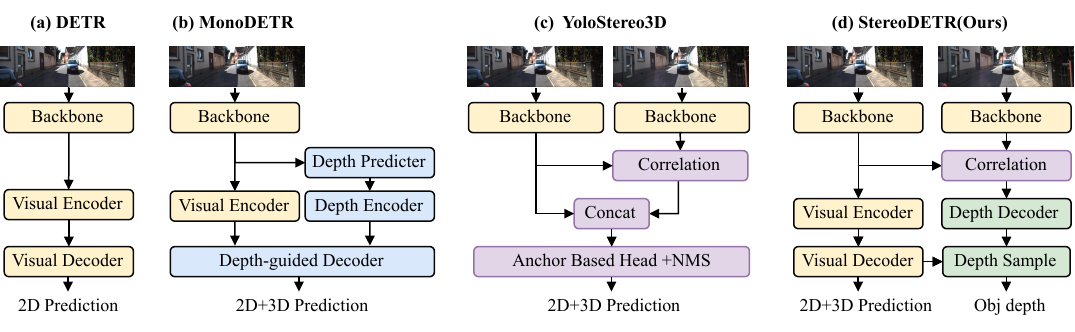}
    \caption{Comparison with existing monocular and stereo architectures. Based on the monocular 2D framework DETR, we introduce a simple extension by incorporating multi-scale correlation computation for depth map estimation, and propose an offset-based depth sampling strategy to address occlusion issues.}
    \label{fig:framework-compare}
\end{figure*}

To resolve the trade-off between accuracy and speed in camera-based methods, we revisit the design of stereo architectures. Specifically, we build on the baseline structure of monocular 2D DETR\cite{Deformable_DETR} and extend it for 3D prediction, aiming to maintain architectural simplicity and avoid excessive computational overhead. We propose the StereoDETR framework, which consists of two branches: a 2D DETR main branch and a stereo-based depth decoder branch. The 2D DETR main branch follows the Deformable-DETR\cite{Deformable_DETR} design and predicts all information except depth, including 2D box, orientation, 3D size, 3D center, and depth sampling point. The stereo-based depth branch efficiently estimates a coarse depth map using a lightweight disparity feature extraction module. These two branches are coupled by a simple sampling strategy. To optimize both inference speed and parameter efficiency, StereoDETR adopts three main simplification strategies in its design: a more lightweight decoder architecture, efficient computation of disparity features, and decoupling of depth prediction from other tasks. 

To simplify the main branch structure of DETR, we streamlined MonoDETR\cite{monodetr_iccv23} and MonoDGP\cite{Monodgp_cvpr25} to a structure similar to DeformableDETR\cite{Deformable_DETR} as the main branch of DETR. We removed the depth encoder and replaced the depth-guided decoder structure with a simple Visual Decoder. Clearly, this simplified branch can serve as a monocular 3D object detection algorithm, with performance potentially lower than MonoDETR and MonoDGP. However, we found that this performance gap primarily arises from depth prediction. When replacing all depth predictions with ground truth values, this gap narrows, and in some cases, this streamlined design shows superior performance. Binocular disparity features can precisely provide depth predictions, allowing us to leverage the computational resources saved by the above simplification process to design a lightweight object-level depth prediction based on disparity. Inspired by Light-Stereo\cite{lightstereo}, the disparity-based depth decoder is designed with a multi-scale disparity feature extraction module. It constructs depth features by aggregating correlation volumes computed from stereo image pairs at multiple scales. These fused features are then fed into a two-layer upsampling convolutional network to predict a target-level depth map such as MonoDETR\cite{monodetr_iccv23}. Unlike MonoDETR, which predicts a 1/16 scale depth map, our method predicts a depth map at 1/4 of the original image resolution. The higher resolution increases the tolerance of the depth sampling process and alleviates occlusion issues. The same upsampling structure is also used to predict a disparity map for auxiliary supervision during training as YOLOStereo3D\cite{YOLOStereo3D_2021}. During inference, only the depth map is predicted.

Our DETR main branch predicts all 3D bounding box attributes except for object depth, while the depth map prediction branch estimates object-level depth. In existing methods, features from the depth prediction branch can be fused into the main branch using cross attention\cite{monodetr_iccv23, Ssd-monodetr_TIV23, Monodgp_cvpr25} or channel concatenation\cite{YOLOStereo3D_2021,TS3D_TITS24,DSC3D_TCSVT25} mechanisms. These representative fusion strategies are illustrated in Fig.~\ref{fig:framework-compare}. Cross-attention approaches require additional depth encoders and attention layers, while channel concatenation methods must balance the proportion of 2D and depth features, often leading to increased channel dimensions. In contrast to these feature-level fusion strategies, we adopt a post-fusion sampling strategy that directly samples from the predicted depth map. However, using the 2D or 3D center points as sampling locations introduces a new issue: occlusion at the object center. To mitigate this, we shift the sampling center to the visible region's center and define the sampling point as an offset from the original center. The main branch box regression head is extended with extra channels to predict this offset. This sampling method is simple and efficient.

We summarize the innovations of this paper as follows.
\begin{itemize}
    \item We propose a novel camera-based stereo 3D object detection algorithm named StereoDETR. To the best of our knowledge, StereoDETR is the first Transformer-based real-time end-to-end 3D detection algorithm.
    \item On public benchmarks, StereoDETR achieves localization performance comparable to non-real-time methods on moderate and hard difficulty levels, and sets new state-of-the-art results for cyclist and pedestrian categories.
    \item StereoDETR breaks the long-standing trade-off between speed and accuracy in stereo and monocular methods, becoming the first stereo 3D detection framework to comprehensively surpass monocular algorithms in speed, with inference time reduced to 17.6 ms.
\end{itemize}

\section{Related Work}
\label{sec:related}

The pure vision-based 3D object detection from a forward-facing perspective can be categorized into two main approaches: monocular and stereo. Monocular methods are low-cost and fast, while stereo methods offer higher accuracy at the expense of speed. A comparison of their accuracy and speed performance are shown in Fig. \ref{fig:ap_time}. In the following, we review the development trends and core challenges of both approaches.

\paragraph{Monocular 3D Object Detection} Compared to 2D object detection, 3D detection tasks require additional predictions of the distance, 3D scale, and orientation of objects. This functionality can be achieved by adding an external regression head to traditional 2D object detection. 
From an architectural perspective, the design of monocular 3D object detection frameworks has evolved from early CNN-based models to more recent Transformer-based architectures.
Models such as SMOKE\cite{smoke_ICCV20}, GUPNet\cite{GUPnet_iccv21}, GUPNet++\cite{GPUnet++_PAMI24}, DID-M3D\cite{Did-m3d-eccv22}, MonoDDE\cite{MonoDDE_cvpr22}, and MonoFlex\cite{MonoFLex_cvpr21} are developed base the CenterNet\cite{Centernet_iccv19}, while FOCS3D\cite{Fcos3d_ICCV21} is built on the FCOS\cite{Fcos-iccv19} framework. MonoDTR\cite{Monodtr_cvpr22} introduces a Transformer-based encoder module but still relies on non-maximum suppression (NMS) for post-processing. In contrast, methods such as MonoDETR\cite{monodetr_iccv23}, SSD-MonoDETR\cite{Ssd-monodetr_TIV23}, and MonoDGP\cite{Monodgp_cvpr25} adopt a fully end-to-end Transformer-based design that eliminates the need for NMS.

One of the key challenges in monocular 3D object detection lies in accurate depth estimation. MonoDDE\cite{MonoDDE_cvpr22} explores multiple strategies for depth estimation fusion, including direct depth regression, scale-aware estimation, and keypoint-based depth prediction. GUPNet\cite{GUPnet_iccv21} introduces uncertainty modeling in both height and depth estimation to enhance the training process. DID-M3D\cite{Did-m3d-eccv22} decomposes depth into three components visual depth, attribute depth, and instance depth, These depth components are independently predicted and aggregated in final. MonoFlex\cite{MonoFLex_cvpr21} investigates an uncertainty-guided aggregation scheme for keypoint-based depth estimation. MonoDTR\cite{Monodtr_cvpr22} incorporates a depth map prediction branch and leverages point cloud information as supervision to facilitate depth feature learning. MonoDETR\cite{monodetr_iccv23} proposes a sparser, object-level depth map prediction branch to guide the network in learning coarse-grained depth maps, and introduces a depth-guided decoding strategy to fuse 2D and depth features effectively. Instead of directly regressing object depth, MonoDGP\cite{Monodgp_cvpr25} decomposes depth estimation into height prediction and height error prediction, aiming to improve robustness. 
To address the occlusion problem, SSD-MonoDETR\cite{Ssd-monodetr_TIV23} introduces Supervised Scale-aware Deformable Attention to enhance the receptive field of queries.
MoVis\cite{MoVis_TIP25} introduces a method that utilizes the depth hierarchy among objects to enhance the depth estimation of occluded targets.

\paragraph{Stereo 3D Object Detection}
The core of binocular 3D object detection lies in how to utilize the disparity information between the two views. Similarly to monocular 3D detection, stereo-based methods have also evolved from early CNN-based monocular 2D detectors. Stereo R-CNN\cite{Stereo-R-CNN-2019} and TLNet\cite{TLNet_CVPR19} are built on the Faster R-CNN framework, while SIDE\cite{SIDE_WACV22}, Stereo CenterNet\cite{StereoCenter_NC22}, and RTS3D\cite{Rts3d_aaai21} are developed based on CenterNet\cite{Centernet_iccv19}. Approaches such as YOLOStereo3D\cite{YOLOStereo3D_2021} and DSC3D\cite{DSC3D_TCSVT25} adopt designs inspired by the YOLO architecture. The first Transformer-based stereo method, TS3D\cite{TS3D_TITS24}, introduces a Transformer-driven disparity-guided positional encoder and decoder structure, but still retains the NMS module for post-processing. In contrast, our work fully adopts the end-to-end DETR framework, eliminating the need for NMS altogether.

Unlike the challenge of inaccurate depth estimation in monocular 3D detection, the primary difficulty in stereo-based methods lies in the inefficiency of depth prediction. RT3DStereo\cite{RT3DStereo_2019} and RT3D-GMP\cite{RT3D-GMP-ITSC20} perform global disparity computation on the original image to estimate scene-wide depth, which is computationally expensive. To accelerate disparity matching, two main strategies have emerged: local object-level matching and low-resolution feature matching. Methods such as Stereo R-CNN\cite{Stereo-R-CNN-2019}, TLNet\cite{TLNet_CVPR19}, and SIDE\cite{SIDE_WACV22} adopt ROI-level disparity matching and fusion strategies, while FGAS R-CNN\cite{FGAS_AEI23} performs instance-level matching based on segmentation masks. On the other hand, YOLOStereo3D\cite{YOLOStereo3D_2021}, DSC3D\cite{DSC3D_TCSVT25}, and TS3D\cite{TS3D_TITS24} conduct fast disparity matching in feature maps at $\frac{1}{4}$ to $\frac{1}{16}$ the original resolution.

Stereo-based methods adopt various strategies for fusing 2D and stereo features. For example, TLNet\cite{TLNet_CVPR19} proposes using disparity to compute a weighted sum of left and right 2D features. SIDE\cite{SIDE_WACV22} and IDA-3D\cite{IDA3d_CVPR20} introduce 2D and depth prediction fusion at the ROI level. Anchor-level fusion strategies are adopted in YOLOStereo3D\cite{YOLOStereo3D_2021}, TS3D\cite{TS3D_TITS24}, and DSC3D\cite{DSC3D_TCSVT25}. 
YOLOStereo3D\cite{YOLOStereo3D_2021} introduces the use of Ghost modules to balance the channel dimensions between 2D and stereo depth features. TS3D\cite{TS3D_TITS24} proposes a multi-scale disparity pyramid fusion strategy. DSC3D\cite{DSC3D_TCSVT25} enhances the adaptability of the receptive field by deformable convolutions. 

Compared to existing stereo methods, our approach employs a more lightweight multi-scale depth feature extraction module, along with a simpler yet computationally efficient multi-task fusion strategy for box and depth prediction.

\section{Method}
\label{sec:metho}

\begin{figure*}[ht] % 双栏图环境
    \centering
    \includegraphics[width=0.8\linewidth]{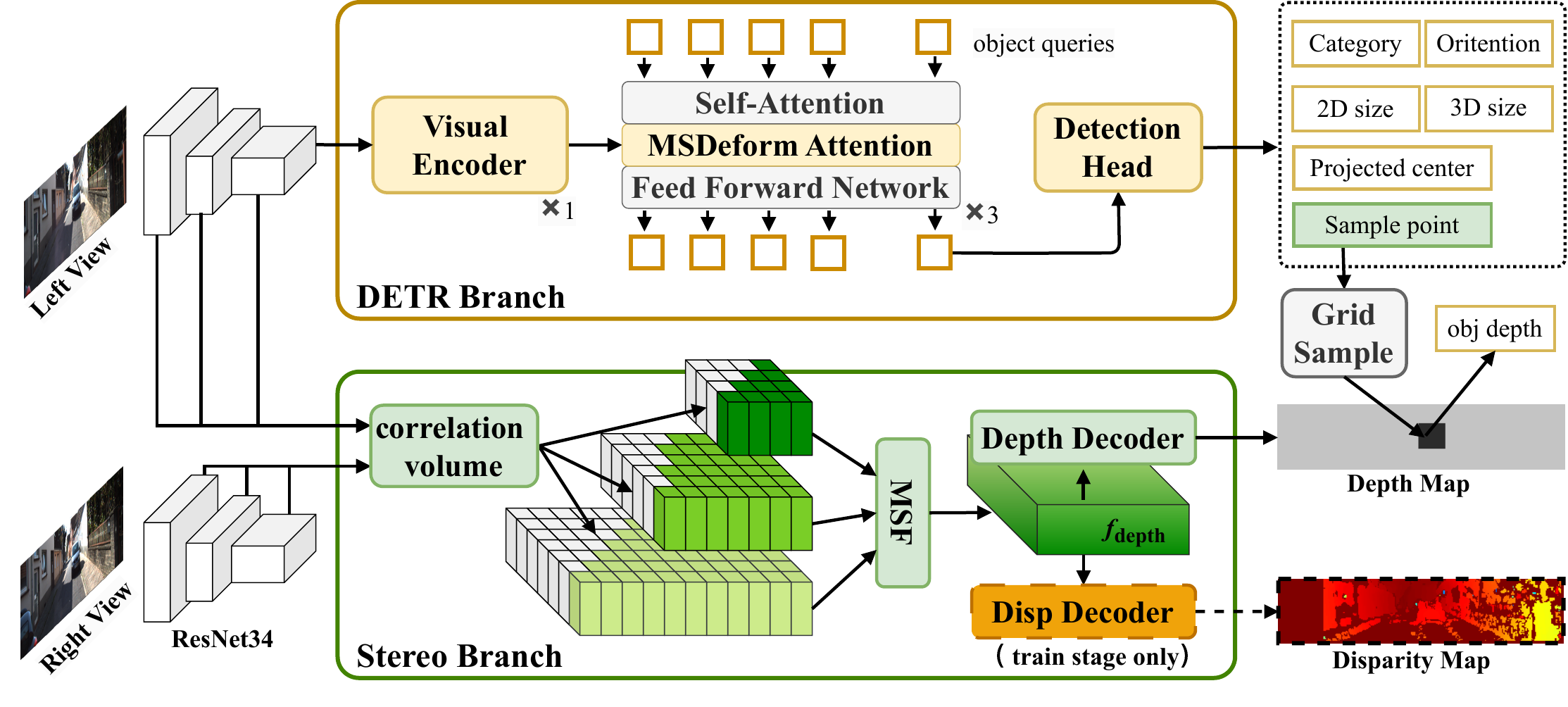}
    \caption{Overview of StereoDETR. The framework consists of two branches: DETR branch and stereo branch. The DETR branch includes a visual encoder-decoder and detection heads to predict object class, orientation, size, 3D center, and depth sampling points. The stereo branch computes a correlation volume, applies Multi-Scale Fusion (MSF), and predicts the depth map and disparity map. The two branches are connected via a Grid Sampling module.
}
    \label{fig:pipeline}
\end{figure*}

As shown in Fig.~\ref{fig:pipeline}, the overall framework of StereoDETR comprises two primary branches: a DETR main branch and a stereo branch. The DETR branch processes features from the left view and is responsible for predicting all attributes except object depth. The stereo branch takes both left and right features, computes cross-view correlation volume, performs multi-scale feature fusion, and utilizes upsampling decoders to estimate a global depth map and a disparity map.

\subsection{Rethinking the Design from Monocular to Stereo 3D Object Detection}

Starting from monocular 2D object detection, we revisit the evolution of detection algorithms through two key stages: from 2D to 3D, and from monocular to stereo. As an initial step, we construct a simple monocular 3D detector by extending a 2D DETR framework, where multiple task heads for 3D prediction are added on end of the decoder. This design is equivalent to adding a depth prediction head to the DETR main branch in our StereoDETR framework. We denote this variant as DETR Branch*. The loss functions and training strategies follow those of MonoDETR\cite{monodetr_iccv23} and MonoDGP\cite{Monodgp_cvpr25}. The upper part of Table~\ref{depth decouple} presents a performance comparison between our DETR Branch*, monocular baselines MonoDETR\cite{monodetr_iccv23} and MonoDGP\cite{Monodgp_cvpr25}, and the stereo-based method YOLOStereo3D\cite{YOLOStereo3D_2021}. It is clear that such a straightforward multi-task extension falls behind existing baseline methods in terms of overall performance. This result also demonstrates the effectiveness of the depth map prediction branches in MonoDETR\cite{monodetr_iccv23} and MonoDGP\cite{Monodgp_cvpr25}.

\begin{table}[b]
\centering
\caption{ $AP_{3D}$ of CAR on KITTI CAR Validation Set with Different Depth Predictions.}
\label{depth decouple}
\begin{tabular}{c|c|c c c}
\toprule
Mothod     &depth mode &Easy  & Mod  & Hard\\ \midrule
YOLOStereo3D\cite{YOLOStereo3D_2021} & prediction  &\textbf{72.06}  &46.58  &35.53\\ 
MonoDETR\cite{monodetr_iccv23}   &prediction  &29.38 &20.63&17.30\\
MonoDGP\cite{Monodgp_cvpr25}    & prediction  & 30.99& 22.50& 19.98\\
DETR Branch*(Ours)  &prediction  &20.39  &16.91  &14.73\\
\midrule

MonoDETR\cite{monodetr_iccv23}   &ground truth&68.38&49.93&38.55\\ 
MonoDGP\cite{Monodgp_cvpr25}    & ground truth &69.46& 52.82& 41.44 \\
DETR Branch*(Ours) &ground truth&66.63 &\textbf{56.76} &\textbf{45.30}\\

\bottomrule

\end{tabular}
\end{table}

However, when switching from monocular to stereo input, depth prediction becomes inherently easier due to the availability of disparity information. This raises a natural question: what is the potential performance of 3D detection under accurate depth estimation?
To explore this, we assume an ideal depth prediction scenario by replacing the predicted depth with ground truth value. The corresponding accuracy results for monocular methods are shown in the lower part of Table~\ref{depth decouple}. In this settings, the simplified DETR Branch* significantly narrows the performance gap with MonoDGP\cite{Monodgp_cvpr25} on the Easy difficulty and even surpasses MonoDGP\cite{Monodgp_cvpr25} on the Moderate and Hard difficulty. This suggests that, given accurate depth estimation, a simple structure can still offer competitive advantages. 
Based on this observation, we extend the DETR main branch by incorporating a depth map prediction branch and a sampling module, aiming to achieve end-to-end stereo 3D object detection with high accuracy and computational efficiency.

\subsection{DETR Main Branch}
\paragraph{Visual Features}

Given the input dual-view images are denoted as \(I_l \in \mathbb{R}^{H \times W \times 3} \) and \(I_r \in \mathbb{R}^{H \times W \times 3} \), \( H\) and \(W\) represent the height and width of the original input images, respectively. The multiscale features obtained by the backbone are denoted as \(f_{\frac{1}{4}} \in \mathbb{R}^{\frac{H}{4} \times \frac{W}{4} \times 64}\),  \(f_{\frac{1}{8}} \in \mathbb{R}^{\frac{H}{8} \times \frac{W}{8} \times 128}\), and  \(f_{\frac{1}{16}} \in \mathbb{R}^{\frac{H}{16} \times \frac{W}{16} \times 256}\). In our final design, we employ Resnet34 as the backbone.

\paragraph{Visual Encoder} For the three-scale features obtained from feature extraction, we expand their channels to 256, flatten them, and then concatenate them, which are then entered into the transformer encoder. The encoder consists of self-attention layers and feedforward layers. To reduce the size of the network, we only use a single encoder layer.

\paragraph{Visual Decoder} The decoder part is designed based on the overall structure of MonoDETR\cite{monodetr_iccv23} and has been simplified. The cross-attention structure involving depth has been removed, whereas the inter-query self-attention, visual cross-attention, and FFN (Feed-Forward Network) have been retained.
At this point, the decoder structure degenerates into a simple DETR decoder structure\cite{Deformable_DETR}, with only additional prediction channels added for predicting the 3D center projection points, rotation angles, 3D scales, and depth map sampling locations. We will introduce the role of 3D sampling points and the method of obtaining pseudo-labels in the following sections.

\subsection{Stereo to Depth Branch}
\paragraph{Stereo Features} 

The process of obtaining stereo features mainly includes two parts: the calculation of the correlation volume and the fusion of multi-scale features. Taking the calculation process of features \( f_{\frac{1}{4},l} \) and \( f_{\frac{1}{4},r} \) at one of the scales as an example, the feature at position \( (i, j) \) in the left view feature is multiplied by the feature at position \( (i-d, j) \) in the right view feature, and the mean is calculated along the channel dimension, the cost value at a scale of \(\frac{1}{4}\), denoted as \( CV_{\frac{1}{4}} \in \mathbb{R}^{\frac{H}{4} \times \frac{W}{4} \times D_1} \) is obtained. 

\begin{equation}
CV_{\frac{1}{4}}(i, j, d)=\frac{1}{64} \sum_{c=1}^{64} f_{\frac{1}{4},l}(i, j) \cdot f_{\frac{1}{4}, r}(i-d, j).
\end{equation}

Similarly, the cost volume are obtained on scales of \(\frac{1}{8}\) and \(\frac{1}{16}\) , denoted as \( CV_{\frac{1}{8}} \in \mathbb{R}^{\frac{H}{8} \times \frac{W}{8} \times D_2} \) and \( CV_{\frac{1}{16}} \in \mathbb{R}^{\frac{H}{16} \times \frac{W}{16} \times D_3} \).

\begin{figure}
    \centering
    \includegraphics[width=0.85\linewidth]{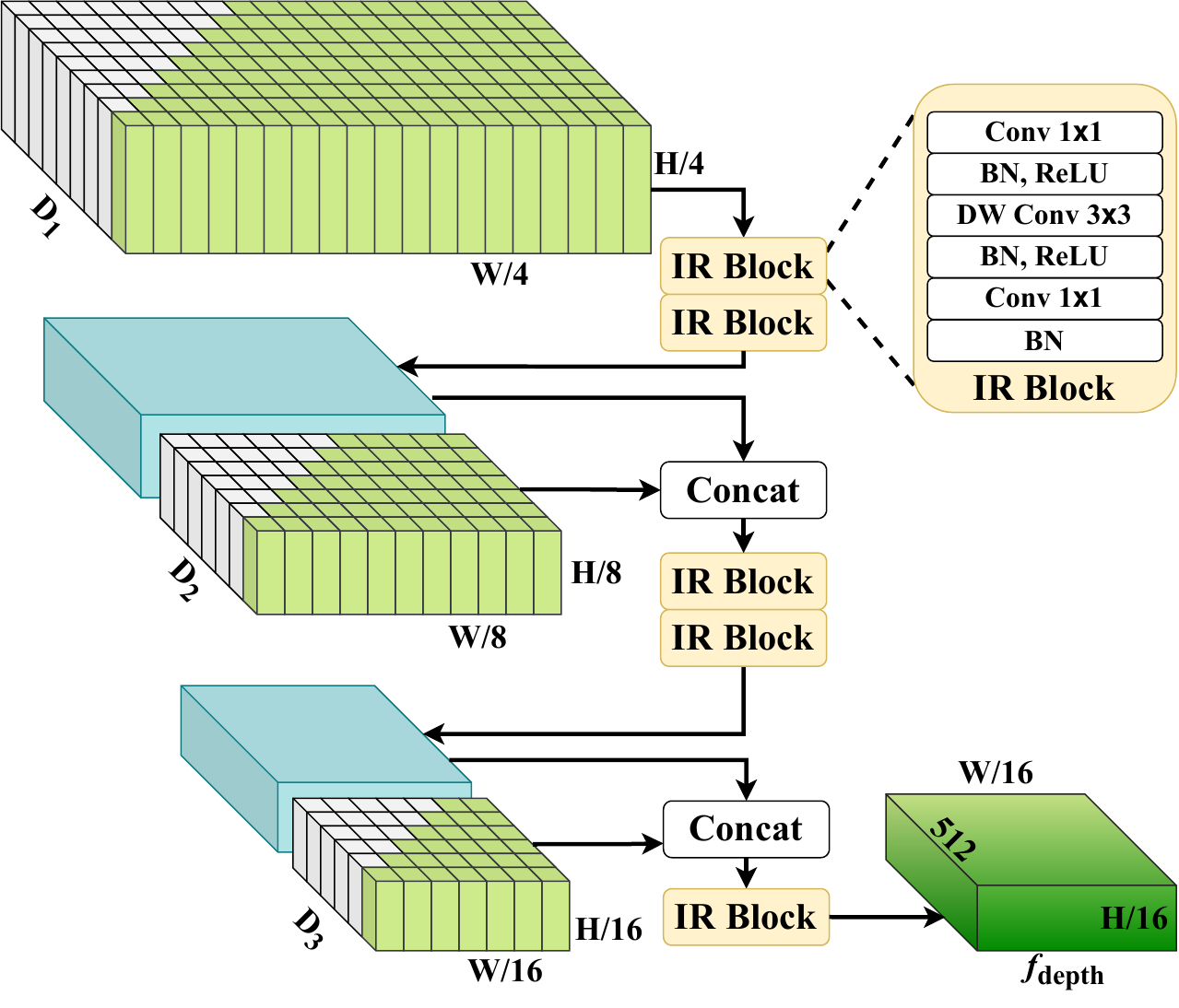}
    \caption{Multi-Scale Fusion module: transforms multi-scale correlation volumes into depth features.}
    \label{fig4:stereo feature}
\end{figure}

\begin{figure}
    \centering
    \includegraphics[width=0.6\linewidth]{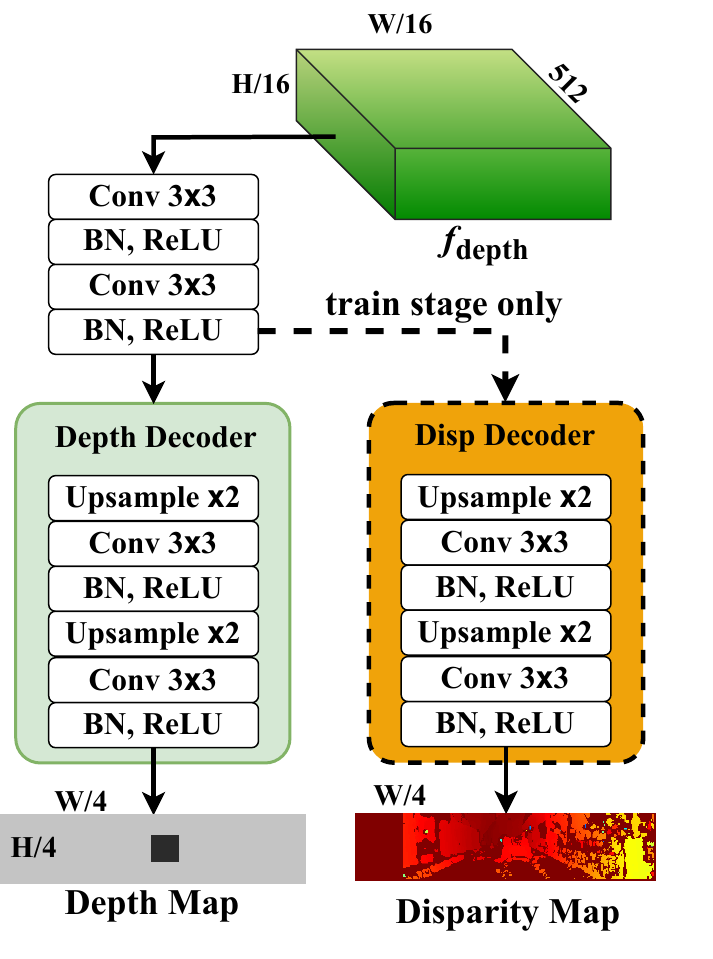}
    \caption{Depth map decoder and disparity decoder modules}
    \label{fig5: depth disp}
\end{figure}

\paragraph{Multi Scale Fusion}
We follow the aggregation strategy in LightStereo\cite{lightstereo} and fuse the multi-scale correlation volumes, as shown in Fig.~\ref{fig4:stereo feature}. The shallow correlation volumes are down-sampled through inverted residual blocks and concatenated with the deep correlation volume.
The final depth feature map obtained is \( f_{depth} \in \mathbb{R}^{\frac{H}{16} \times \frac{W}{16} \times 512} \).

\paragraph{Depth map and Disparity Map Decoder}
Unlike using pixel-level depth maps provided by LiDAR point clouds for supervision, this paper continues with two coarse and low-cost depth supervision methods in the pure vision-based approach: object-level depth map  \( M_{depth} \in \mathbb{R}^{\frac{H}{4} \times \frac{W}{4} \times 80}\) and disparity map  \( M_{disp} \in \mathbb{R}^{\frac{H}{4} \times \frac{W}{4} \times 96} \), Here, 80 and 96 denote the number of output channels, which are kept consistent with the reference baselines\cite{monodetr_iccv23,YOLOStereo3D_2021}.
As shown in Fig.~\ref{fig5: depth disp}, The depth map decoder and the disparity decoder are composed of two upsample and convolution layers. 
In MonoDETR\cite{monodetr_iccv23}, the predicted depth map resolution is $\frac{1}{16}$ of original input. In this paper, a higher resolution of $\frac{1}{4}$ is adopted to mitigate the impact on occluded objects.
The disparity map prediction is used only as an auxiliary supervision during training and can be discarded during the test inference phase.

\subsection{Object Depth Sample}

\begin{figure}
    \centering
    \includegraphics[width=0.8\linewidth]{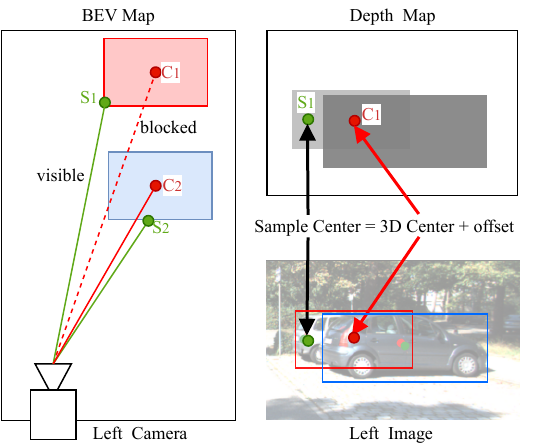}
    \caption{Depth sampling strategy designed to address occlusions challenge.}
    \label{fig6: depth sample}
\end{figure}

As shown in Fig.~\ref{fig6: depth sample}, the 3D center point $C_1$ of the distant white car is occluded by the nearby gray vehicle. If this center point is used for sampling on the depth map $M_{depth}$, it will output the incorrect depth value of the nearby vehicle, rather than the true value of the blocked one.
To address this issue, we designed learnable sampling points to sample from visible region. To supervise the predicted sampling points, a very simple calculation strategy was employed. The Depth Map is drawn as a rectangle using the Depth value of the object as the color, within the region of the 2D box, and overlapping areas retain the value of the closer object. For the visible regions of occluded objects, which are usually rectangular or L-shaped, the sub-rectangle with the largest vertical span is calculated, and the center point of this sub-rectangle is used as the ground truth for supervising the sampling points. In the sampling phase, we directly use PyTorch's \texttt{grid\_sample} function to perform sampling. This operation is differentiable and supports gradient backpropagation. It has also been widely used in learnable geometric transformations, such as in Spatial Transformer Networks (STN).

\subsection{Loss Functions}
The loss function primarily consists of two components: object-level loss and global-level loss. The object-level loss is composed of classification loss, 2D loss, and 3D loss. The 2D loss includes the 2D center point, 2D size, projected point of the 3D center, and depth sampling points. The 3D loss comprises the depth distance of the object, 3D scale estimation, and rotation angle. The global-level loss includes object-level depth map loss and disparity reconstruction loss, which are designed with reference to MonoDETR\cite{monodetr_iccv23} and YOLOStereo3D\cite{YOLOStereo3D_2021}, respectively. The total loss function is as follows:
\begin{equation}
\mathcal{L}_{\text{total}}=\mathcal{L}_{\text{obj}}+ \mathcal{L}_{\text{global}}
\end{equation}
\begin{equation}
\mathcal{L}_{\text{obj}}=\frac{1}{N} \cdot \sum_{n=1}^{N}\left(\mathcal{L}_{\text{cls}}+\mathcal{L}_{\text{2d}}+\mathcal{L}_{\text{3d}}\right)
\end{equation}
\begin{equation}
\mathcal{L}_{\text{global}}= \mathcal{L}_{\text{depth}} + \mathcal{L}_{\text{disp}}
\end{equation}

\paragraph{Bipartite Matching}
We first apply the hungarian matching algorithm to perform one-to-one bipartite matching between the $N$ predicted boxes and $N_{\text{gt}}$ ground-truth boxes, resulting in $N_{\text{gt}}$ matched pairs. Following MonoDGP\cite{Monodgp_cvpr25} and MonoDETR\cite{monodetr_iccv23}, we compute the matching cost using only 2D information, where the 2D matching cost consists of classification cost and box regression cost.

\begin{equation}
\mathcal{C}_{\text{match}} = \mathcal{C}_{\text{cls}} + \mathcal{C}_{\text{box2D}}
\end{equation}

\begin{equation}
\mathcal{C}_{\text{cls}} (i, j) =
\begin{cases}
  -\alpha \cdot (1 - y_i)^\gamma \cdot \log(y_i), & \text{if } \hat{y}_j = 1 \\
  (1 - \alpha) \cdot y_i^\gamma \cdot \log(1 - y_i), & \text{otherwise.}
\end{cases}
\end{equation}

The 2D bounding box regression cost is computed using the L1 distance between the predicted and ground-truth box coordinates:
\begin{equation}
\mathcal{C}_{\text{box2D}} (i, j) = 
\lvert b_i - \hat{b}_j \rvert + \text{GIoU}(b_i, \hat{b}_j)
\end{equation}
where $\hat{b}_j \in \mathbb{R}^4$ is ground truth 2D box, and $b_i$ is prediction.

\paragraph{Object Level Loss}

For each matched pair, we compute the classification loss, 2D bounding box regression loss, 3D size and center regression loss, depth sampling point supervision, and object depth loss. The regression and classification losses follow the design of the baseline network MonoDETR\cite{monodetr_iccv23}, where Focal Loss~\cite{Focal_loss} is used for classification and L1 loss is applied for regression tasks. In particular, we retain the uncertainty-based loss\cite{uncertainties_nips2017} for object depth prediction.
\begin{equation}
\mathcal{L}_{obj} = 
\mathcal{L}_{\text{cls}}
  + \mathcal{L}_{\text{reg}}
  + \mathcal{L}^{\text{obj}}_{\text{depth}}
\end{equation}

\begin{equation}
\mathcal{L}^{obj}_{\text{depth}} = \sqrt{2} \cdot e^{-\sigma} \cdot \lvert d - \hat{d} \rvert + \sigma
\end{equation}
Here, $\sigma$ is predicted by the DETR branch decoder head, representing the predicted uncertainty of the model.

\paragraph{Global Level Loss}

To enhance global consistency, we introduce two auxiliary supervision losses: the disparity reconstruction loss and the depth map supervision loss. The ground-truth disparity map is generated using a traditional block matching algorithm. For depth map supervision, each 2D bounding box is filled with its corresponding object 3D center depth value to construct a pseudo-depth map. The disparity loss adopts the same formulation as YOLOStereo3D~\cite{YOLOStereo3D_2021}, while the depth map supervision follows the design in MonoDETR~\cite{monodetr_iccv23}.
\section{Experiments}
\label{sec:exp}

\begin{table*}
\centering
\caption{3D Object Detection of Car Category on The KITTI Test Server.}
\label{kitti_test_server_3D_bev_2D_car}
% {@{}c|ccc@{}}
\begin{tabular}{@{}c|c|ccc|ccc|ccc|c@{}}
\toprule &  &\multicolumn{3}{c|}{$AP_{3D}$} & \multicolumn{3}{c|}{ $AP_{BEV}$} & \multicolumn{3}{c|}{$AP_{2D}$}& \multirow[t]{2}{*}{Runtime} \\
\multirow[t]{2}{*}{Method} &\multirow[t]{2}{*}{Train Data}    & Easy  & Mod   & Hard  & Easy  & Mod   & Hard  & Easy  & Mod   & Hard   & (ms)\\ \midrule
MonoGeo\cite{MonoGeo_21}   & Monocular     & 18.85 & 13.81 & 11.52 & 25.86 & 18.99 & 16.19 & - & - & -  & 50 \\
MonoFlex\cite{MonoFLex_cvpr21} & Monocular & 19.94 & 13.89 & 12.07 & 28.23 & 19.75 & 16.89 & 96.01&	91.02 &	83.38   & 30 \\
DEVIANT\cite{Deviant_eccv22}  & Monocular  & 21.88 & 14.46 & 11.89 & 29.65 & 20.44 & 17.43 & 94.42 & 86.64 & 76.69  & 40 \\
MonoDETR\cite{monodetr_iccv23} & Monocular & 23.65 & 15.92 & 12.99 & 32.08 & 21.44 & 17.85 & 93.99 & 86.17 & 76.19   & 20 \\
MonoJSG\cite{Monojsg_cvpr22}  & Monocular  & 24.69 & 16.14 & 13.64 & 32.59 & 21.26 & 18.18 & - & - & -  & 42 \\
MonoEdge\cite{monoedge_WACV23}  & Monocular& 21.08 & 14.47 & 12.73 & 28.80 & 20.35 & 17.57 & - & - & -  & 37 \\

GUPNet++\cite{GPUnet++_PAMI24}  & Monocular
                               & 24.99 & 16.48 & 14.58 & - & - & - & - & - & -  & 34 \\   
SSD-MonoDETR\cite{Ssd-monodetr_TIV23}  & Monocular
                               & 24.52 & 17.88 & 15.69 & 33.59 & 24.35 & 21.98 & - & - & -  & 21 \\ 
MonoDGP\cite{Monodgp_cvpr25}  & Monocular
                               & 26.35 & 18.72 & 15.97 & 35.24 & 25.23 & 22.02 & - & - & -  &  22\\ 
                               \midrule
DSGN\cite{DSGN_CVPR20}  & Stereo+LiDAR        &73.50  &52.18 &45.14 &82.90 &65.05 &56.60 &95.53 &86.43 &78.75 & 670 \\ 
StereoDistill & Stereo+LiDAR&81.66 &66.39&57.39&89.03 &78.59 &69.34&97.61&	93.43 &87.71 &400\\
OC-Stereo\cite{OC-Stereo-ICRA20}  & Stereo+Depth+Mask & 55.15&	37.60 &	30.25 &68.89 &	51.47 &	42.97 &87.39 &	74.60 &	62.56 & 350 \\
DMF\cite{DMF_TITS22}  &Stereo+Depth&77.55 &	67.33 &	62.44  &84.64  &80.29  &76.05&89.50&	85.49 &	82.52 &200\\
ESGN \cite{ESGN_TCSVT22}  & Stereo+LiDAR&65.80&46.39&38.42&78.10&	58.12&49.28&93.07 &80.58&70.68 & 60\\ \midrule

SIDE\cite{SIDE_WACV22}   & Stereo       & 47.69 & 30.82 &25.68 & - & - & -& - & - & - & 260 \\

StereoRCNN\cite{Stereo-R-CNN-2019} & Stereo
                               & 47.58 & 30.23 & 23.72 & 61.92 & 41.31 & 33.42 & 93.98 & 85.98 & 71.25 & 200 \\

DSC3D\cite{DSC3D_TCSVT25}   & Stereo    & \textbf{66.46} & \textbf{42.54} & \underline{34.04} & \underline{74.56} & 51.21 & 42.07 & \underline{96.56} & 88.74 & 76.41 & 110 \\

TLNet\cite{TLNet_CVPR19}   & Stereo     & 7.46  & 4.37  & 3.74  & 13.71 & 7.69  & 6.73  & 76.92 & 63.53 & 54.58  &  100 \\
FGAS RCNN\cite{FGAS_AEI23}  & Stereo & 58.02 & 38.68 & 32.53 & 72.56 & \underline{58.31} & \underline{46.24} & - & - & - & 100 \\

TS3D\cite{TS3D_TITS24}   & Stereo       & 64.61 & \underline{41.29} & 30.68 & 73.34 & 48.59 & 36.98 & 92.39 & 77.23 & 57.28 & 88 \\ 

YoloStereo3D\cite{YOLOStereo3D_2021}  & Stereo
                               & \underline{65.68} & 41.25 & 30.42 & \textbf{76.10} & 50.28 & 36.86 & 94.81 & 82.15 & 62.17 & 80 \\ 

RT3DStereo\cite{RT3DStereo_2019} & Stereo
                               & 29.90 & 23.28 & 18.96 & 58.81 & 46.82 & 38.38 & 56.53 & 45.81 & 37.63 &  79 \\

RT3D-GMP\cite{RT3D-GMP-ITSC20}  & Stereo& 45.79 & 38.76 & 30.00 & 69.14 & \textbf{59.00} & 45.49 & 62.41 & 51.95 & 39.14 & 60\\

Stereo CenterNet\cite{StereoCenter_NC22} & Stereo
                               & 49.94 & 31.30 & 25.62 & 62.97 & 42.12 & 35.37& \textbf{96.61} & \underline{91.27} & \underline{83.50} & 43\\ 

RTS3D\cite{RTD3D_AAAI2021}  & Stereo    & 58.51 & 37.38 & 31.12 & 72.17 & 51.79 & 43.19 & - & - & - & \underline{39}\\ 

    \midrule
StereoDETR(Ours)     & Stereo           &59.45 &41.17 &\textbf{35.13} &72.77 & 54.53 & \textbf{46.41} &96.39 & \textbf{93.45} & \textbf{83.67} &\textbf{17.6} \\
 \bottomrule
\end{tabular}
\end{table*}

\subsection{Settings}

\paragraph{Dataset}
We choose to use the KITTI\cite{KITTI} dataset for training and validation.
The KITTI dataset consists of 7,481 annotated training samples and 7,518 unlabeled test samples publicly available. For the official server evaluation, we trained a model using all 7,481 training samples containing annotations for cars, pedestrians, and cyclists, and submitted the test results. In the ablation study phase, following \cite{Chen_Split} and YOLOStereo3D\cite{YOLOStereo3D_2021}, we divided the 7,481 training samples into 3,712 training samples and 3,769 evaluation samples. The training process exclusively utilized RGB images from left and right views, camera parameters, and 3D bounding box annotations, without introducing additional auxiliary supervision such as point clouds or ground plane annotations.

\paragraph{Evaluation Metric}    
In both online testing and offline evaluation, we assess the performance of cars, pedestrians, and cyclists. We use 3D Average Precision $AP_{3D}$, Bird’s Eye View Average Precision ($AP_{BEV}$), and 2D Average Precision ($AP_{2D}$) as metric. Objects are categorized into different difficulty levels, such as easy moderate and hard, based on occlusion levels and the size of their 2D bounding boxes. For the test results of online server, we directly report the returned scores. For offline evaluation and ablation studies, we adopt recall@40 instead of recall@11.

\paragraph{Implementation details}
We follow the design of YOLOStereo3D\cite{YOLOStereo3D_2021} in the stereo vision approach, using ResNet34 as the backbone network for feature extraction and removing the top 100 sky regions with less informative content. 
The input image resolution to the model is 288$\times$1280. 
The training and evaluation are conducted on a single RTX 4090 GPU. The initial learning rate is set to 0.0002, which decays by a factor of 0.1 at epochs 125 and 165, with a total of 195 training epochs. The batch size is set to 16 during the training phase.

\subsection{Main Results}

\paragraph{Results of Car category on the KITTI test server}
Table~\ref{kitti_test_server_3D_bev_2D_car} reports a comparison between the proposed method and existing state-of-the-art (SOTA) pure vision-based 3D object detection methods on the KITTI test server. The comparison is limited to the Car category. Table~\ref{kitti_test_server_3D_bev_2D_car} is divided into three parts. The first part includes pure monocular 3D object detection methods, which can typically achieve real-time performance, but their $AP_{3D}$ is only about half that of stereo-based approaches. The second part covers stereo methods guided by additional information, although these methods achieve significantly higher accuracy than monocular and purely stereo-based methods, they suffer from low inference speed and cannot meet real-time requirements. The third part presents purely stereo-based training methods. We sort this part by runtime, and it is evident that our method, StereoDETR is the first stereo-based approach to surpass monocular methods in both speed and accuracy. 
Compared to the state-of-the-art fastest real-time stereo solution RTS3D\cite{Rts3d_aaai21} with a latency of 39 ms (25.64 FPS), StereoDETR only requires 17.6 ms (56.82 FPS), achieving a 120\% increase in FPS. Compared with non-real-time methods, StereoDETR lags behind YoloStereo3D\cite{YOLOStereo3D_2021} and its derivatives TS3D\cite{TS3D_TITS24} and DSC3D\cite{DSC3D_TCSVT25} under the Easy level difficulty. However, StereoDETR remains competitive under Moderate and Hard, while offering a four-times speed advantage.

\paragraph{Results of Car catgory on KITTI validation set}
To further compare the performance with existing SOTA methods under different IoU thresholds, Table~\ref{kitti_eval_offline_3D_bev_2D_car} reports the 3D and BEV evaluation results for the Car category, using the commonly adopted train validation split\cite{Chen_Split}. 
In the table, we only report purely stereo methods that use the same data split.
Consistent with the result from the online test set, StereoDETR outperforms existing real-time methods on the \textit{easy} level of the validation set. On the \textit{hard} level, it even surpasses all pure stereo-based approaches, achieving a new state-of-the-art. However, on the \textit{easy} level, it still lags behind some non-real-time methods. The primary reason lies in the design choice to simplify the architecture, where the 3D size is predicted solely based on monocular features. As a result, the 3D estimation capability is inherently inferior to that of methods that utilize feature-level fusion. A similar phenomenon can also be observed in Table~\ref{depth decouple}, where DETR branch still underperforms compared to other methods on the \textit{easy} level, even under the ideal assumption of using ground-truth depth maps.

\begin{table*}
\centering
\caption{3D Object Detection of Car Category on The KITTI Validation Set.}
\label{kitti_eval_offline_3D_bev_2D_car}
% {@{}c|ccc@{}}
\begin{tabular}{@{}c|ccc|ccc|ccc|ccc|c@{}}
\toprule & \multicolumn{3}{c|}{ $AP_{3D}$ (IoU=0.7)} & \multicolumn{3}{c|}{$AP_{BEV}$ (IoU=0.7)} & \multicolumn{3}{c|}{$AP_{3D}$ (IoU=0.5)}& \multicolumn{3}{c|}{$AP_{BEV}$ (IoU=0.5)} & \multirow[t]{2}{*}{Runtime} \\
\multirow[t]{2}{*}{Method}   & Easy  & Mod  & Hard & Easy & Mod  & Hard & Easy & Mod  & Hard & Easy & Mod & Hard & (ms)\\ \midrule
SIDE\cite{SIDE_WACV22}       & 61.22 &44.46 &37.15 &72.75 &53.71 &46.16 &87.70 &69.13 &60.05 &88.35 &76.01 &67.46 &260\\
StereoRCNN\cite{Stereo-R-CNN-2019}
                             & 54.11 &36.69 &31.07 &68.50 &48.30 &41.47 &85.84 &66.28 &57.24 &87.13 &74.11 &58.93 & 200\\
DSC3D\cite{DSC3D_TCSVT25}    & \textbf{73.78} &\textbf{48.90} &\underline{39.85} &\textbf{83.56} &57.74 &48.12 &\textbf{94.92} &77.29 &65.49 &\underline{95.35} &80.07 &68.26 &110\\

TLNet\cite{TLNet_CVPR19}     & 18.15 &14.26 &13.72 &29.22 &21.88 &18.83 &59.51 &43.71 &37.99 &62.46 &45.99 &41.92   & 100 \\
FGAS RCNN\cite{FGAS_AEI23}   & 57.92 &40.84 &34.07 &68.16 &48.26 &47.15 &89.81 &72.46 &64.62 &90.25 &78.77 &62.31 &100\\
IDA3D \cite{IDA3d_CVPR20}    & 54.97 &37.45 &32.23 &70.68 &50.21 &42.93 &87.08 &74.57 &60.01 &88.05 &76.69 &67.29&-\\
YoloStereo3D\cite{YOLOStereo3D_2021}
                             & \underline{72.06} &46.58 &35.53 &\underline{80.69} &55.22 &43.47 &- &- &- &- &- &- &80\\
Stereo CenterNet\cite{StereoCenter_NC22}
                             & 55.25 &41.44 &35.13 &71.26 &53.27 &45.53 &- &- &- &- &- &- &43 \\
RTS3D\cite{RTD3D_AAAI2021}   & 64.76 &46.70 &39.27 &77.50 &\underline{58.65} &\underline{50.14} &90.34 &\underline{79.67} &\underline{70.29} &90.58 &\underline{80.72} &\underline{71.41} &\underline{39}\\
    \midrule
    
StereoDETR(Ours)       & 68.95 &\underline{48.23} &\textbf{41.47} &79.78 &\textbf{58.97} &\textbf{51.69} &\underline{92.92} &\textbf{80.10} &\textbf{72.56} &\textbf{95.36} &\textbf{82.99} &\textbf{75.49} &\textbf{17.6}\\
% StereoDETR(Ours)       & 68.17 &49.06 &41.35 &- &- &- &- &- &- &- &- &- &17.6 \\
% StereoDETR(Ours)       & 68.99 &48.25 &41.47 &- &- &- &- &- &- &- &- &- &17.6\\
 \bottomrule
\end{tabular}
\end{table*}

\paragraph{Results of Pedestrian category on the KITTI test set}
In contrast to large-scale objects like cars, pedestrians present greater detection challenges due to their smaller spatial extent and less distinctive features.
To further demonstrate the performance on more challenging categories, Table~\ref{kitti_test_server_3D_bev_2D_ped} presents a comparison between existing methods and our approach on the Pedestrian from the test server. 
Compared with the CNN-based state-of-the-art method DSC3D\cite{DSC3D_TCSVT25}, StereoDETR achieves performance improvements of 3.58\%, 2.82\%, and 1.94\% on the 3D detection metric under the Easy, Moderate, and Hard levels.
On the BEV metric, the improvements are 4.72\%, 3.32\%, and 3.35\%. 

\begin{table*}
\centering
\caption{3D Object Detection of Pedestrian Category on The KITTI Test Server.}
\label{kitti_test_server_3D_bev_2D_ped}
% {@{}c|ccc@{}}
\begin{tabular}{@{}c|c|ccc|ccc|ccc|c@{}}
\toprule &  &\multicolumn{3}{c|}{$AP_{3D}$} & \multicolumn{3}{c|}{ $AP_{BEV}$} & \multicolumn{3}{c|}{$AP_{2D}$}& \multirow[t]{2}{*}{Runtime} \\
\multirow[t]{2}{*}{Method}  &  \multirow[t]{2}{*}{Train Data}   & Easy  & Mod   & Hard  & Easy  & Mod   & Hard  & Easy  & Mod   & Hard   & (ms)\\ \midrule
MonoFlex\cite{MonoFLex_cvpr21} & Monocular& 9.43 & 6.31 & 5.26 & 10.36 & 7.36 & 6.29 &62.64 &47.58 &43.15 & 30\\
DEVIANT\cite{Deviant_eccv22}  & Monocular & 13.43 & 8.65 & 7.69 & 14.49 & 9.77 & 8.28&74.27 &55.16 &50.21&40 \\ 
MonoDTR \cite{Monodtr_cvpr22}  & Monocular  & 15.33 & 10.18 & 8.61 & 16.66 & 10.59 & 9.00&59.44 &42.86 &38.57 & 40\\
GUPNet++\cite{GPUnet++_PAMI24} & Monocular  & 12.45 & 8.13 & 6.91 & - & - & -&- &- &- & 37\\
MonoDGP\cite{Monodgp_cvpr25} & Monocular  & 15.04 & 9.89 & 8.38 & 15.78 & 10.43 & 8.79 &64.38 &48.95 &44.59 & 22\\
\midrule
DSGN\cite{DSGN_CVPR20} & Stereo+LiDAR& 20.53 & 15.55 & 14.15 &26.61&20.75&18.86 &49.28&39.93 &38.13 & 670\\
StereoDistill\cite{Stereodistill_AAAI23} & Stereo+LiDAR& 44.12 &32.23 & 28.95 &50.79 &37.75 &	34.28&69.00 &55.09 &50.95 &400  \\
OC-Stereo\cite{OC-Stereo-ICRA20} & Stereo+Depth+Mask & 24.48& 17.58& 15.60  & 29.79& 20.80&18.62&43.50 &30.79 &	28.40 & 350\\ 
DMF\cite{DMF_TITS22} &Stereo+Depth &37.21 &	29.77&	27.62 &42.08 &	34.92 &	32.69&52.99&43.43&41.29 &200 \\
ESGN \cite{ESGN_TCSVT22}& Stereo+LiDAR &14.05&10.27&9.02&17.94&13.03&11.54&44.09 &32.60&	29.10 & 60\\
\midrule

RT3DStereo\cite{RT3DStereo_2019} & Stereo& 3.28 & 2.45 & 2.35 &4.72&3.65 & 3.00&41.12&29.30&25.25 &80\\
RT3D-GMP\cite{RT3D-GMP-ITSC20} & Stereo &16.23 & 11.41 & 10.12 &19.92&14.22&12.83&55.56 &39.83&35.18&60\\

YOLOStereo3D\cite{YOLOStereo3D_2021} & Stereo &28.49 & 19.75 & 16.48 &31.01 &20.76&18.41&\underline{56.20}&\underline{41.46} &\underline{37.07} & 80 \\ 
DSC3D\cite{DSC3D_TCSVT25}  & Stereo& \underline{29.60} & \underline{20.43} & \underline{17.92} &\underline{32.25}& \underline{22.57}&\underline{19.01}&-&-&-&110\\ \midrule
StereoDETR(Ours) & Stereo & \textbf{33.18}  &\textbf{23.25} & \textbf{19.86}&\textbf{36.97}&\textbf{25.89} & \textbf{22.26}&\textbf{76.89} &\textbf{59.62}& \textbf{54.58}&\textbf{17.6} \\ 
 \bottomrule
\end{tabular}
\end{table*}

\begin{table}
\centering
\caption{3D Object Detection of Cyclist Category on The KITTI Test Server.}
\label{test_on_cyclist}
\begin{tabular}{c|c c c }
\toprule
       & Easy  & Mod  & Hard \\ 
Method & $3D/BEV$  & $3D/BEV$  & $3D/BEV$ \\ 
\midrule

MonoFlex\cite{MonoFLex_cvpr21} & 4.17  / 4.41  & 2.35 / 2.35 & 2.04 / 2.50 \\ 

DEVIANT\cite{Deviant_eccv22} & 5.05  / 6.42  & 3.13 / 3.97 & 2.59 / 3.51 \\ 
MonoDTR \cite{Monodtr_cvpr22}   &5.05 / 5.84 & 3.27 / 4.11 & 3.19 / 3.48 \\ 
GUPNet++\cite{GPUnet++_PAMI24}    &6.71 / \phantom{0}---& 3.91 / \phantom{0}---& 3.80 / \phantom{0}---\\ 
MonoDGP \cite{Monodgp_cvpr25}   &5.28 / 6.48 & 2.82 / 3.61 & 2.65 / 3.22 \\ 
\midrule
DSGN\cite{DSGN_CVPR20} &27.76 / 31.23&18.17 / 21.04&16.21 / 18.93 \\
ESGN\cite{ESGN_TCSVT22} &13.84 / 15.78 & 7.6 9 / 9.02 & 6.75 / 7.96  \\
StereoDistill\cite{Stereodistill_AAAI23} &63.96 / 69.46 &	44.02 / 48.37 &39.19 / 42.69 \\
OC-Stereo\cite{OC-Stereo-ICRA20} & 29.40 / 32.47& 16.63 / 19.23  & 14.72 / 17.11\\
DMF\cite{DMF_TITS22}  &65.51 / 71.92&51.33 / 57.99& 45.05 / 51.55\\
\midrule
RT3DStereo\cite{RT3DStereo_2019} &5.29 / 7.03& 3.37 / 4.10 & 2.57 / 3.88\\

RT3D-GMP\cite{RT3D-GMP-ITSC20}& 18.31 / 20.59 & 12.99  / 13.92  &  10.63 / 12.74\\

StereoDETR(Ours)   &\textbf{39.09}  /\textbf{ 42.19}  & \textbf{24.29} / \textbf{26.78} & \textbf{20.77} / \textbf{22.74}  \\ 
\bottomrule
\end{tabular}
\end{table}

\paragraph{Results of Cyclist category on the KITTI test set}

In Table~\ref{test_on_cyclist}, we report the 3D and BEV detection performance of existing methods and our proposed method on the \textit{Cyclist} class from the test server. Compared with the \textit{Pedestrian} results in the previous Table~\ref{kitti_test_server_3D_bev_2D_ped}, it is evident that monocular methods\cite{MonoFLex_cvpr21, Deviant_eccv22, Monodtr_cvpr22,GPUnet++_PAMI24} tend to perform worse on the \textit{Cyclist} class. For example, under the \textit{Easy} setting, MonoFlex\cite{MonoFLex_cvpr21} achieves an $AP_{3D}$ of 9.43\% for \textit{Pedestrian} but only 4.17\% for \textit{Cyclist}, while GUPNet++\cite{GPUnet++_PAMI24} achieves 12.45\% and 6.71\% for \textit{Pedestrian} and \textit{Cyclist}. This performance drop is mainly due to the rarity of \textit{Cyclist} samples.
In contrast, stereo-based or depth-supervised methods show the opposite trend. RT3D-GMP\cite{RT3D-GMP-ITSC20} achieves an $AP_{3D}$ of 16.23\% on \textit{Pedestrian} and 18.31\% on \textit{Cyclist}. StereoDETR achieves 33.18\% and 39.09\% on \textit{Pedestrian} and \textit{Cyclist} with easy difficulty, demonstrating superior robustness stereo-based methods. Moreover, our approach outperforms not only purely stereo-based methods such as RT3D-Stereo\cite{RT3DStereo_2019} and RT3D-GMP\cite{RT3D-GMP-ITSC20} across all three difficulty levels on the \textit{Cyclist} class in both 3D and BEV metrics, but also surpasses several LiDAR-supervised approaches, including OC-Stereo\cite{OC-Stereo-ICRA20}, ESGN\cite{ESGN_TCSVT22}, and DSGN\cite{DSGN_CVPR20}.

\subsection{Ablation Studies}
\paragraph{3D Scale Prediction Mode}
In our approach, 3D scale prediction is implemented by increasing the number of prediction channels in the 2D DETR head. Two prediction strategies are considered: directly predicting the absolute 3D scale, or predicting the projected scale in the 2D image view.
As shown in Table~\ref{scale_prediction}, the performance difference between predicting absolute scale and predicting projected scale is minor, with $AP_{3D}$ variations of only 0.81\%, -0.27\%, and 0.98\% on the \textit{Easy}, \textit{Moderate}, and \textit{Hard} levels.

\begin{table}
\centering
\caption{Comparison of different 3D Scale prediction modes ($AP_{3D}$ for Car).}
\label{scale_prediction}
\begin{tabular}{c|c c c}
\toprule
3D Scale mode     &Easy  & Mod  & Hard\\ \midrule
Absolute Scale   &\textbf{68.95} &48.23 &\textbf{41.47}\\
Projected Scale  &68.14 &\textbf{48.50} &40.49\\ 
\bottomrule
\end{tabular}
\end{table}

This is largely due to the fact that the KITTI dataset's training and validation sets are randomly split from the same set of driving sequences, leading to similar vehicle models and viewing angles in both sets. As a result, the network can learn to regress absolute size directly from monocular features, without requiring intermediate projection-based computations using estimated height and depth. This also explains why MonoDGP\cite{Monodgp_cvpr25} is able to estimate the depth by predicting the 3D height of the object rather than directly regressing the depth.

\paragraph{Decoder Ablation Study}
In monocular DETR-based frameworks\cite{monodetr_iccv23,Monodgp_cvpr25}, due to the coupled 2D and 3D prediction head, it is necessary to fuse depth features with 2D features, and the decoder stage requires guidance from depth representations. In contrast, our method adopts a decoupled design, allowing the use of a standard 2D visual decoder for object-level prediction without the depth guidance. Table~\ref{decoder_mode} compares the performance between a depth ware decoder\cite{monodetr_iccv23} and a simple visual decoder. The results show that the simple decoder is more efficient in terms of inference speed. Moreover, by avoiding error propagation from imperfect depth map estimation, it achieves slightly better accuracy in 3D prediction.

\begin{table}
\centering
\caption{Ablation study on decoder design($AP_{3D}$ for Car).}
\label{decoder_mode}
\begin{tabular}{c|c c c | c }
\toprule
Decoder mode     &Easy  & Mod  & Hard & Runtime$ \downarrow$\\ \midrule
Depth aware decoder\cite{monodetr_iccv23}  &68.18 &47.73 &40.93 & 20.4 ms\\
Visual decoder   &\textbf{68.95} &\textbf{48.23} &\textbf{41.47} &\textbf{17.6} ms\\ 
\bottomrule
\end{tabular}
\end{table}

\begin{figure*}[ht] 
    \centering
    \includegraphics[width=1\linewidth]{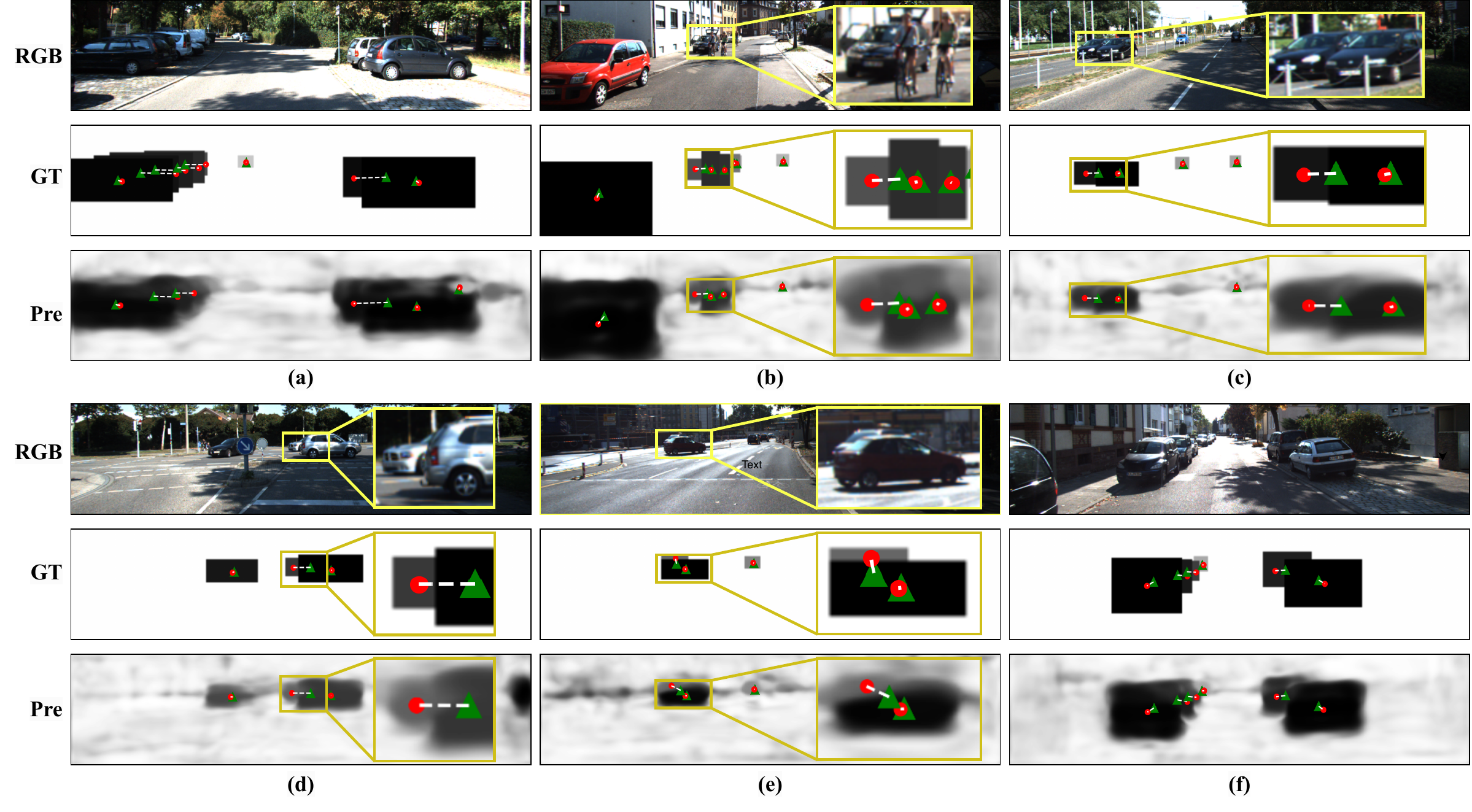}
    \caption{Visualization results of predicted depth maps, 3D object centers, and non-occluded sampling points.}
    \label{fig7:depth_sample}
\end{figure*}

\paragraph{Depth sampling strategy}

In Table~\ref{depth_sample_mode}, we compare the impact of different depth prediction modes and sampling methods on StereoDETR. High-resolution depth map prediction enhances fault tolerance for depth sampling. When using 3D centers as sampling points, increasing resolution from 1/16 to 1/4 yields improvements of 3.55\%, 1.63\%, and 2.52\% for Easy, Moderate, and Hard levels. Building upon high-resolution(1/4) inputs, offset sampling strategy mitigates incorrect depth sampling caused by occlusion, achieving gains of 3.33\%, 3.59\%, and 5.51\% across the three difficulty levels, with particularly notable improvements in Moderate and Hard. Crucially, our supervision sample point ground truth can be generated from 3D bounding box annotations, eliminating the need for additional manual labeling.

The visualization results of depth prediction and sample points are shown in Fig.~\ref{fig7:depth_sample}, where samples from six validation sequences are illustrated. \textbf{GT} denotes the ground truth, \textbf{Pre} denotes the predicted results. Blocks with different gray levels represent object depths, green triangles indicate the 3D center points, red circles mark the visible sampling points. It can be observed that the predicted depth distribution and sampling points closely match the ground truth. Compared with the potentially occluded 3D centers, the adaptively offset sampling points fall within the correct non-occluded regions.

\begin{table}
\centering
\caption{ Ablation results on KITTI validation set for different depth sampling strategies and depth map scales($AP_{3D}$ for Car).}
\label{depth_sample_mode}
\begin{tabular}{c|c|c c c}
\toprule
depth map scale    & depth sample pos &Easy  & Mod  & Hard\\ \midrule
1/16        & 3D center &62.07 &43.01 &33.44\\ 
1/16        & offset    &66.48 &47.35 &40.74\\ \midrule
1/4        & 3D center  &65.62 &44.64 &35.96 \\
1/4        & offset   &\textbf{68.95} &\textbf{48.23} &\textbf{41.47}  \\
 \bottomrule

\end{tabular}
\end{table}

\paragraph{Auxiliary Supervision Loss Ablation Study}

In addition to the object-level supervision losses, we also introduce two global-level auxiliary supervision signals: depth map loss and disparity map loss. The depth maps are generated from the ground truth 3D center position, while the ground truth disparity maps are obtained using a block matching algorithm. Table~\ref{depth map and disp map loss} presents the effect of these auxiliary supervision losses on AP$_{3D}$. The introduction of depth map supervision improves AP$_{3D}$ by 1.32\%, 0.42\%, and 0.22\% on the \textit{Easy}, \textit{Moderate}, and \textit{Hard} levels. Adding disparity supervision provides further gains of 2.30\%, 0.98\%, and 1.19\% on the corresponding difficulty levels.

\begin{table}
\centering
\caption{Ablation study on the effect of depth map and disparity map supervision losses($AP_{3D}$ for Car).}
\label{depth map and disp map loss}
\begin{tabular}{c c|c c c }
\toprule
depth map loss & disparity map loss     &Easy  & Mod  & Hard \\ \midrule
$\times$     &  $\times$       &65.33 &46.83 &40.06\\
% $\times$     &  $\checkmark$   &65.79 &47.18 & 40.49\\
$\checkmark$ & $\times$        &66.65 &47.25 & 40.28\\
$\checkmark$ & $\checkmark$    &\textbf{68.95} &\textbf{48.23} & \textbf{41.47}\\ 
\bottomrule
\end{tabular}
\end{table}

\paragraph{Lightweight and acceleration settings}

To enable single-GPU training and reduce memory consumption, we follow the backbone and preprocessing crop top strategy used in YOLOStereo3D\cite{YOLOStereo3D_2021}. Table~\ref{tricks_for_lightweight} shows the differences in $AP_{3D}$ and inference latency on the car class under different settings. In terms of $AP_{3D}$ performance, the accuracy differences among the three experimental setups are marginal. Using a smaller version of ResNet saves approximately 1.1\,ms in inference time, while cropping the top 100 pixels reduces latency by about 0.5\,ms. The speed advantage of  StereoDETR mainly comes from its simplified architecture, which avoids the additional latency introduced by the fusion of deep features and 2D texture features.

\begin{figure*} 
    \centering
    \includegraphics[width=1\linewidth]{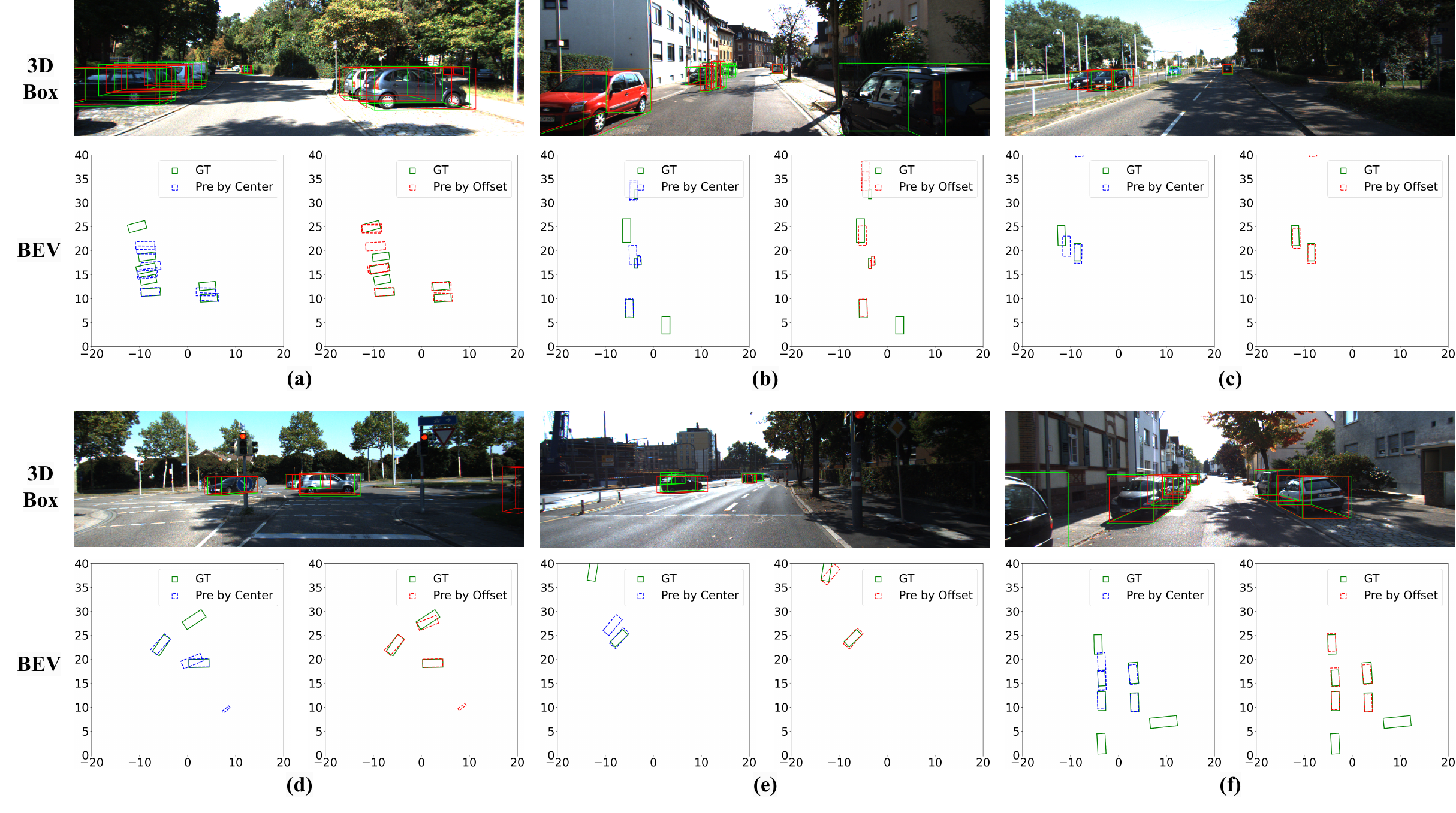}
    \caption{Visualization results of the predicted 3D bounding boxes and their corresponding representations in the Bird's-Eye View.}
    \label{fig:bev_3d}
\end{figure*}

\begin{table}[h]
\centering
\caption{Ablation study of backbone and input cropping strategies for acceleration.}
\label{tricks_for_lightweight}
\begin{tabular}{c c|c c c | c }
\toprule
Backbone & CropTop     &Easy  & Mod  & Hard & Runtime\\ \midrule
Resnet34 & $\checkmark$  &\textbf{68.95} &48.23 &41.47 &\textbf{17.6} ms \\ 
Resnet50 & $\checkmark$  &67.14 &\textbf{49.55} &\textbf{41.87} &18.8 ms\\
Resnet34 & $\times$   &68.06 &48.07 &41.50 &18.3 ms\\

\bottomrule
\end{tabular}
\end{table}

Table~\ref{Gflops_params} compares the computational complexity and parameter size of existing monocular and stereo-based methods. The proposed StereoDETR achieves a lower computational cost of 59.80\,GFLOPs, compared to 62.96\,GFLOPs of the monocular MonoDETR\cite{monodetr_iccv23}, with a corresponding inference speedup of 2.4\,ms. We adopt a lightweight backbone with only 8.2\,M parameters as YOLOStereo3D\cite{YOLOStereo3D_2021}. Furthermore, we avoid the use of depth encoders and complex decoder designs. As a result, the total number of parameters in StereoDETR is only 18.4\,M, which corresponds to 48.8\% of that in MonoDETR\cite{monodetr_iccv23}, and merely 17.1\% of YOLOStereo3D\cite{YOLOStereo3D_2021}.

\begin{table}
\centering
\caption{Comparison of computational load and parameter count.}
\label{Gflops_params}
\resizebox{\columnwidth}{!}{%
\begin{tabular}{c c c  c c c}
\toprule
\multirow{2}{*}[-0.5ex]{\centering \textbf{Method}} & 
\multirow{2}{*}[-0.5ex]{\centering \textbf{time(ms)}} &
\multirow{2}{*}[-0.5ex]{\centering \textbf{GFLOPs$\downarrow$}} & 
\multicolumn{3}{c}{\textbf{Parameters$\downarrow$ (M)}} \\
\cmidrule(lr){4-6}
& & & \textbf{Total} & \textbf{Backbone} & \textbf{Other} \\
\midrule
MonoDETR\cite{monodetr_iccv23}    & 20&  62.96  & 37.7 & 23.5 & 14.2 \\
MonoDGP\cite{Monodgp_cvpr25}     & 22&  71.79  & 42.2 & 23.5 & 18.7 \\
YOLOStereo3D\cite{YOLOStereo3D_2021} &80& 177.82  &107.6 &  \textbf{8.2} & 99.4 \\ \midrule
StereoDETR(Ours)   &\textbf{17.6} &  \textbf{59.80} & \textbf{18.4} &  \textbf{8.2} & \textbf{10.2} \\
\bottomrule
\end{tabular}
}
\end{table}

\subsection{Qualitative Results}

As shown in Fig.~\ref{fig:bev_3d}, we visualize the predicted 3D bounding boxes for six samples, which are consistent with those in Fig.~\ref{fig7:depth_sample}. In the camera view, green boxes denote ground truth and red boxes indicate predictions. In the bird’s-eye view, green boxes represent ground truth, blue dashed boxes correspond to predictions based on 3D center sampling, and red dashed boxes are results based on offset sampling. In subfigure (b), two cyclists occlude a car, and the center-based sampling strategy leads to an overly close prediction. This issue is alleviated by the offset-based sampling. Occluded vehicles in subfigures (c), (d), (e), and (f) also show improved predictions. However, in subfigure (a), where multiple vehicles are densely occluded on the left side, the proposed method still requires improvement. 
\section{Conclusion}
\label{sec:conc}

To achieve a balance between computational efficiency and accuracy in stereo vision, we rethink and redesign a concise DETR-based stereo 3D detection framework named StereoDETR. By employing lightweight disparity-aware feature extraction and a task-decoupling strategy, we reduce the reliance on fusion modules and complex architectures. For the first time, our stereo-based method surpasses monocular 3D detectors in inference speed, reaching 56.8 FPS. A supervised depth sampling strategy is introduced to unify depth estimation and object detection, leading to improved accuracy on hard, occluded targets. On the KITTI test set, our method achieves new state-of-the-art performance on the challenging pedestrian and cyclist categories. Thanks to the simplicity and efficiency of our design, the computational cost of stereo 3D detection is significantly reduced, improving real-time performance and laying a solid foundation for future research on open-vocabulary 3D detection in open world.

\begin{IEEEbiography}[{\includegraphics[width=1in,height=1.25in,clip,keepaspectratio]{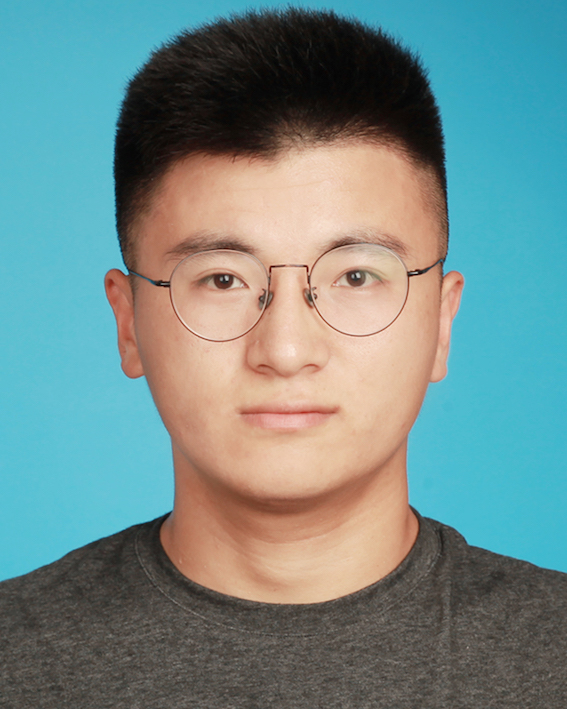}}]{Shiyi Mu}
received the M.Eng. Degree from the
School of Communication and Information Engineering, Shanghai University, China, in 2022. He
is currently pursuing the Ph.D. degree with the
information and communication engineering, Shanghai University, China. His research interests include
deep learning for computer vision, optical character recognition, 3D object detection, and anomaly detection.
\end{IEEEbiography}
% \vspace{5pt}
\begin{IEEEbiography}[{\includegraphics[width=1in,height=1.25in,clip,keepaspectratio]{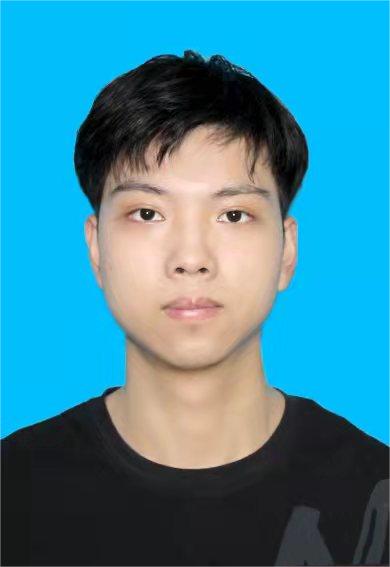}}]{Zichong Gu}
 received the B.Eng. degree from the Department of Communication Engineering, Shanghai University, Shanghai, China, in 2023, where he is currently pursuing the M.Eng. degree with the School of Communication and Information Engineering. His research interests include autonomous driving, depth estimation and open-vocabulary detection.
\end{IEEEbiography}
\vspace{5pt}
\begin{IEEEbiography}[{\includegraphics[width=1in,height=1.25in,clip,keepaspectratio]{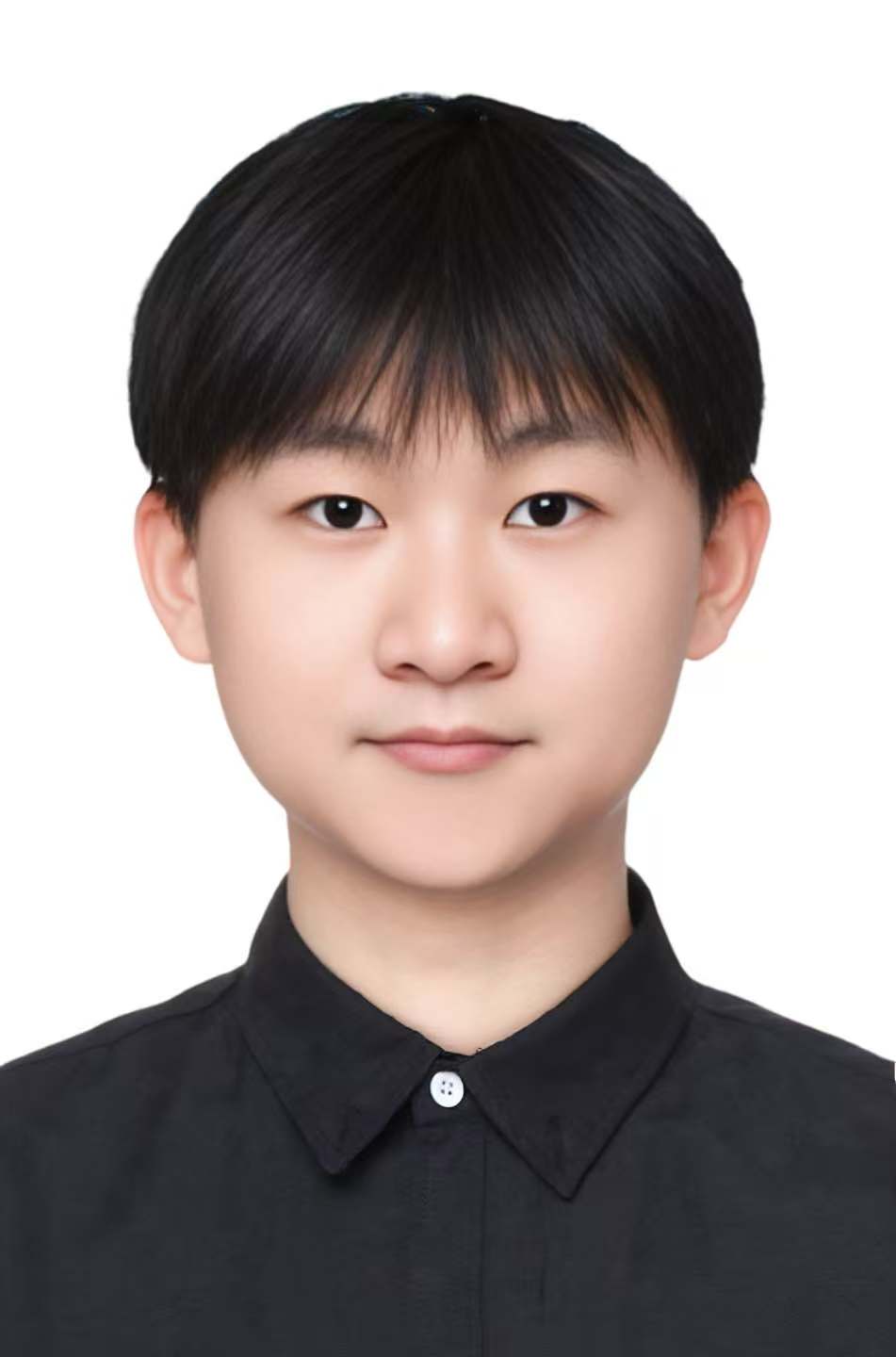}}]{Zhiqi Ai}
 received his Bachelor's degree in Communications Engineering from Shanghai University in 2021 and his Master's degree in Engineering from the same university in 2024. He is currently pursuing a Ph.D. in Information and Communication Engineering at Shanghai University, China. His research interests include personalized speech recognition, speech synthesis, and multimodal large language models.
\end{IEEEbiography}
\vspace{5pt}
\begin{IEEEbiography}[{\includegraphics[width=1in,height=1.25in,clip,keepaspectratio]{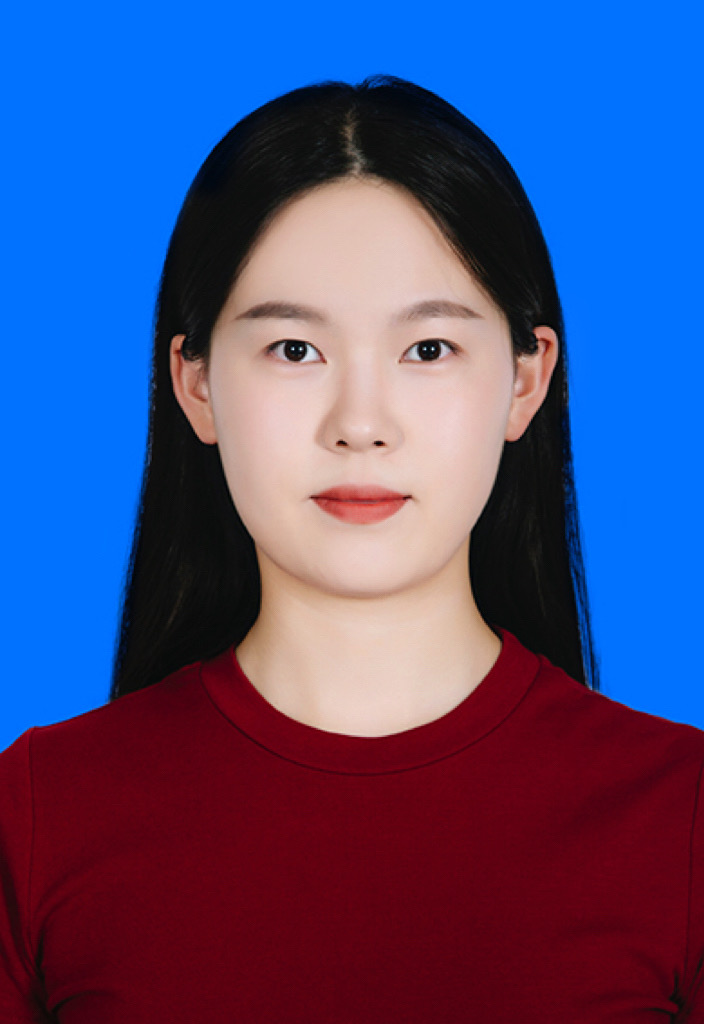}}]{Anqi Liu} received the B.Eng. degree from the School of Communication and Information Engineering, Shanghai University, China, in 2023. She is currently pursuing the M.Eng. degree with the information and communication engineering, Shanghai University, China.  Her research interests include autonomous driving simulation, AI ISP, data generation, and 3D object detection.
\end{IEEEbiography}
\vspace{5pt}
\begin{IEEEbiography}[{\includegraphics[width=1in,height=1.25in,clip,keepaspectratio]{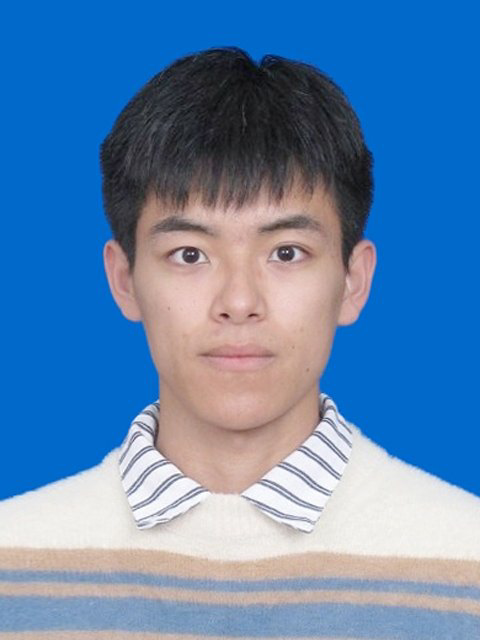}}]{Yilin Gao} received the B.Eng. degree from the Department of Communication Engineering, Shanghai University, Shanghai, China, in 2021. He is currently pursuing the Ph.D. degree in information and communication engineering at Shanghai University, China. His research directions cover OCR, Object Detection, Autonomous Driving, AIGC, and Embodied Intelligence.\end{IEEEbiography}
% \vspace{5pt}
\begin{IEEEbiography}[{\includegraphics[width=1in,height=1.25in,clip,keepaspectratio]{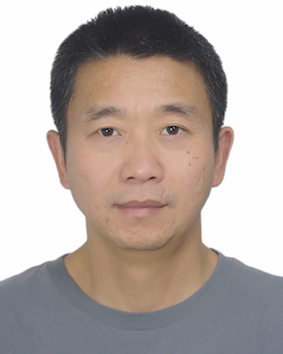}}]{Shugong Xu}
(Fellow, IEEE) graduated from Wuhan University, China, in 1990, and received his Master degree in Pattern Recognition and Intelligent Control from Huazhong University of Science and Technology (HUST), China, in 1993, and Ph.D. degree in EE from HUST in 1996. He is now a full professor and the AVP-R at Xi’an Jiaotong-Liverpool University. He was the center Director and Intel Principal Investigator of the Intel Collaborative Research Institute for Mobile Networking and Computing (ICRI-MNC), prior to December 2016 when he joined Shanghai University as a full professor. Before joining Intel Labs in September 2013, he was a research director and principal scientist at the Communication Technologies Laboratory, Huawei Technologies. He was also the Chief Scientist and PI for the China National 863 project on End-to-End Energy Efficient Networks. Shugong was one of the co-founders of the Green Touch consortium together with Bell Labs etc. Prior to joining Huawei in 2008, he was with Sharp Laboratories of America as a senior research scientist. Before that, he conducted research as research fellow in City College of New York, Michigan State University and Tsinghua University. Dr. Xu was elevated to IEEE Fellow in 2015. He is also the winner of the 2017 Award for Advances in Communication from IEEE Communications Society. His current research interests include machine learning and pattern recognition,  intelligent machine, as well as ISAC in wireless communication systems.
\end{IEEEbiography}

\end{document}